\documentclass[11pt]{article}

\usepackage[final]{acl}

\usepackage{times}
\usepackage{latexsym}
\usepackage{multirow}
\usepackage{amsmath}
\usepackage{amssymb}
\usepackage{multirow}
\usepackage{makecell}
\usepackage{tabularx}
\usepackage{booktabs}
\usepackage{graphicx}
\usepackage{fontawesome5}
\usepackage{subcaption}
\usepackage[table]{xcolor}

\usepackage[T1]{fontenc}

\usepackage[utf8]{inputenc}

\usepackage{microtype}

\usepackage{inconsolata}

\title{Teaching Language Models to Check Grounded Claim Factuality with Human Test-Taking Strategies}

\author{Yuxuan Ye \and Raul Santos-Rodriguez \and Edwin Simpson \\
  Intelligent Systems Laboratory \\
  University of Bristol \\
  United Kingdom \\
  \texttt{\{yuxuan.ye, enrsr, edwin.simpson\}@bristol.ac.uk} \\}

\begin{document}
\maketitle
\begin{abstract}
Grounded claim factuality checking is important for large language model (LLM) applications such as retrieval-augmented generation, as it helps users assess the correctness of generated outputs.
Existing metrics using entailment classifiers require dataset-specific threshold tuning, while LLM-based approaches often use direct prompting, which underutilises the reasoning capabilities of LLMs.
We address this by formulating grounded claim factuality checking as a true/false reading comprehension task and prompting LLMs with explicit test-taking strategies for efficient reasoning.
Our method reduces token usage by over 80\% compared to unguided open-ended reasoning, and achieves competitive performance to more expensive alternatives across two factuality benchmarks, setting a new state of the art on one.
To further reduce inference cost, we train small language models (SLMs) to replace LLMs in the checking pipeline.
Using supervised fine-tuning (SFT) and a self-revision mechanism, the SLMs learn to improve their factuality judgements. 
Experimental results show that the resulting SLMs perform on par with strong baselines,
combining low inference costs with generating supporting rationales to improve interpretability. \footnote{Code is available at \url{https://github.com/Haruhi07/Test-Taking}}
\end{abstract}

\section{Introduction}
Large language models (LLMs) are applied to a wide variety of generation tasks, including summarisation and question answering, that require the generated content to be consistent with grounding documents.
However, LLM outputs do not always satisfy this condition and are prone to making up statements that are not supported by the given sources \citep{zhang-2024-how,mckenna-etal-2023-sources}.
As a result, detecting such hallucinations is critical for conditioned generation tasks to ensure the trustworthiness and reliability of the outputs.

Recent work often regards factuality evaluation as a textual entailment task \citep{lei-etal-2025-factcg,zha-etal-2023-alignscore,laban-etal-2022-summac}.
This line of research mainly employs a classifier to predict an entailment score, which has the advantage of being lightweight and computationally efficient.
However, the entailment scores require a dataset-specific threshold to convert them into explicit \textit{True} and \textit{False} predictions.
This limitation motivates us to design an evaluation metric that outputs factuality judgements and generalises across datasets without threshold tuning.

Given a suitably designed prompt, LLMs can directly decide whether a claim is grounded in a source document \citep{xu-etal-2024-identifying,luo-2023-chatgpt}.
Supported by recent advances in understanding and reasoning capabilities
\citep{qwen3,deepseek-r1,dubey-2024-llama}, 
this ability has led to the \emph{LLM-as-judge} methodology for factuality checking.
A major challenge for this methodology lies in designing prompts that fully leverage the LLM's reasoning ability \citep{xu-etal-2024-identifying,liu-etal-2023-g,luo-2023-chatgpt}. 
An additional benefit of LLM-as-judge is that, with an appropriate prompt, LLMs can generate rationales for judgements, which provide information for identifying errors in the claim and locating evidence in the source document.

However, LLMs are computationally intensive when deployed at scale.
A potential alternative is to adapt small language models (SLMs), which have recently been shown to handle certain reasoning tasks when trained properly \citep{qwen3}. 
Their lightweight architecture and ability to generate reasoned judgements without threshold tuning 
motivate further exploration of SLMs for claim factuality checking, including strategies to adapt and train them for this purpose.

\begin{figure*}[htbp]
    \centering
    \includegraphics[width=0.95\linewidth]{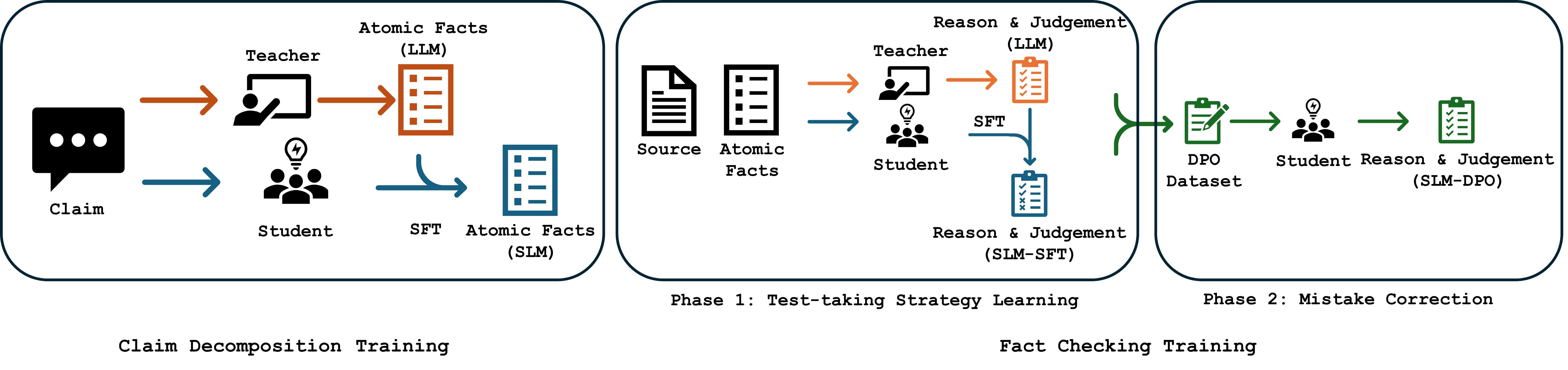}
    \caption{Overview of SLM training. \textbf{Claim decomposition training} (left) utilises the output from an LLM (teacher) as the reference output in SFT for the SLM (student). \textbf{Fact checking training, phase 1:} (middle)
    LLM (teacher) outputs are used to train an SLM (student) to follow a test-taking strategy, in which the model separately assesses each atomic fact. \textbf{Phase 2:} (right) We further refine the student model by pairing its incorrect outputs with correct outputs from the teacher model, then fine-tuning the student's reasoning output via DPO.
    }
    \label{fig:overview}
\end{figure*}

In our work, we design a claim checking pipeline that incorporates LLM-as-judge into a human test-taking strategy.
We first decompose each claim into atomic facts to simplify the subsequent fact checking step.
In the second step, motivated by the similarity between fact checking and True/False reading exercises in language proficiency tests, we design a prompt that leverages test-taking strategies commonly used in such assessments.
Unlike previous approaches, in which the model directly outputs its judgement or reasoning without guidance, we require the model to follow the strategies to reason and make judgements.
Experimental results demonstrate the effectiveness of this pipeline, achieving competitive performance to more expensive alternatives across two factuality benchmarks, including a new state of the art on one of them.

We then propose a training method to adapt a general purpose SLM for claim factuality evaluation.
Given the limited generalisation capacity of SLMs, we train separate models for claim decomposition and fact checking.
Using our LLM-as-judge strategy, we employ an LLM as a teacher for both tasks, using supervised fine-tuning (SFT) to train the student SLM. 
For the fact checking step, the student model further revises its rationales through direct preference optimisation (DPO) 
\citep{rafailov2023direct} using the reference output from the teacher model.
In this way, the student model first learns the test-taking strategy and then strengthens this ability by correcting its own mistakes, resulting in higher fact checking accuracy than relying solely on SFT.
The overall training process is shown in Figure \ref{fig:overview}.

Our contributions are three-fold:
\begin{itemize}
    \item Inspired by test-taking strategies, we design a prompt for grounded claim checking that simulates how humans process claims in assessments. Our approach achieves state of the art on one benchmark, while generating rationales with judgements.
    \item We propose a two-phase training procedure and demonstrate that the resulting SLMs achieve performance approaching that of LLMs, despite being a fraction of the size.
    \item To the best of our knowledge, this is the first work to apply SLMs to claim factuality evaluation by generating reasoned judgements, demonstrating their potential to evaluate factuality in a reasoning pipeline and provide interpretable judgements efficiently.
\end{itemize}

\section{Related Work}
\subsection{Factuality Evaluation Metrics}
Several widely used factuality metrics assess a claim's factual consistency by measuring its semantic similarity to the grounding document \citep{zhang-2019-bertscore,ye-2024-using}, but they fail to capture small textual differences that alter factual content while maintaining high semantic similarity.

The resemblance between factuality evaluation and textual entailment has motivated the use of entailment classifiers for factuality \citep{lei-etal-2025-factcg,tang-etal-2024-minicheck,zha-etal-2023-alignscore,laban-etal-2022-summac}, which has gradually become the mainstream paradigm.
Entailment classifiers detect whether a hypothesis is supported by a premise at sentence-level.
In claim factuality checking, they often need to truncate or chunk the source document, leading to information loss that decreases performance.

Another line of work \citep{fabbri-etal-2022-qafacteval,deutsch-etal-2021-towards,wang-etal-2020-asking} regards factuality evaluation as a question answering (QA) problem, measuring a claim's factuality by comparing answers generated when conditioning on the claim versus the grounding document.
While effective, these QA-based methods often rely on complex multi-stage pipelines.
In contrast, our work adopts the reading comprehension concept for claim factuality evaluation while simplifying the pipeline into two concise steps.

\subsection{LLM in Factuality Evaluation}
LLMs are both generators of content to be evaluated and powerful tools for developing evaluation metrics.
One application of LLMs is synthesising training data to improve factuality metrics \citep{lei-etal-2025-factcg,tang-etal-2024-minicheck,feng-etal-2024-teaching-small}.
These approaches often augment training data by generating adversarial or challenging examples.

Another use case of LLMs leverages their understanding and reasoning ability to directly perform factuality evaluation, commonly referred as LLM-as-judge.
Some studies \citep{luo-2023-chatgpt,gekhman-etal-2023-trueteacher} evaluated the performance of directly prompting an LLM to decide the correctness of a claim.
While these works also investigate the benefits of LLM reasoning, the models are left to reason their judgements without explicit guidance.
Other approaches \citep{xu-etal-2024-identifying,liu-etal-2023-g} explore incorporating pre-defined criteria into the prompt to guide judgements.
However, the provided criteria are limited to error type definitions or scoring standards, rather than an explicit step-by-step reasoning plan for factuality evaluation.

In this paper, we utilise LLMs in both ways. 
Specifically, we 1) design a prompt that adapts a general purpose LLM into an effective factuality evaluation metric, and 2) use an LLM to generate training data for fine-tuning an SLM to perform factuality judgement. 
Unlike prior work, we prompt the LLM with a structured strategy that explicitly guides the evaluation step by step, leading to more systematic and interpretable judgements.  

\subsection{SLM Fine-tuning}
Recent advances in LLMs have also accelerated progress in SLMs.
Distillation from strong teacher models provides an efficient way to transfer capabilities, improving SLM performance on tasks such as reasoning and mathematics \citep{qwen3,deepseek-r1}. 
A common approach is to generate rationales using LLMs and fine-tune student SLMs via SFT \citep{feng-etal-2024-teaching-small,gekhman-etal-2023-trueteacher,jiang-etal-2023-lion}.
However, these approaches rely solely on SFT to imitate the output from the teacher model.
Previous works \citep{qwen3,deepseek-r1} mention that performance can be improved by further leveraging the synthesised data with reinforcement learning following SFT.
In this vein, we apply DPO after SFT, enabling the student model to improve its reasoning accuracy by learning from its own mistakes, thereby aligning better with the teacher model.

\section{Method}
In this section, we first reformulate claim factuality checking as a reading comprehension task.
Then we introduce a pipeline that adapts a general purpose LLM for factuality evaluation, 
taking inspiration from human test-taking strategies.
Finally, we present our approach to fine-tune an SLM for use in the evaluation pipeline.

\subsection{Problem Reformulation}
\citet{lei-etal-2025-factcg} defines that, given a grounding document $doc$ and a claim $c$, $c$ is grounded in $doc$ if the statement \textit{"According to $doc$, $c$ is true"} is generally affirmed by a generic reader.
This definition reformulates grounded claim checking into a True/False reading comprehension problem.
Therefore, this paper tackles claim factuality evaluation by utilising generative language models to solve reading comprehension tasks, leveraging a strategy that a human examinee could apply in such tasks.

\subsection{LLM-based Evaluation Pipeline}
\label{sec:LLM}
Recent LLMs achieve high scores on human language proficiency tests, including the reading comprehension problem formulated above \citep{achiam-2023-gpt}.
Noting that human test-takers often use multi-step reasoning strategies to improve their performance on reading comprehension tests \citep{yapp-2023-effects,al2021empirical}, 
we incorporate such a strategy into prompt design, enabling LLMs to process grounded fact checking efficiently and systematically. 
By guiding the model’s reasoning process, our approach aims to improve both accuracy and inference efficiency by shortening the model's reasoning outputs.

\begin{figure}[tbp]
    \centering
    \includegraphics[width=0.9\linewidth]{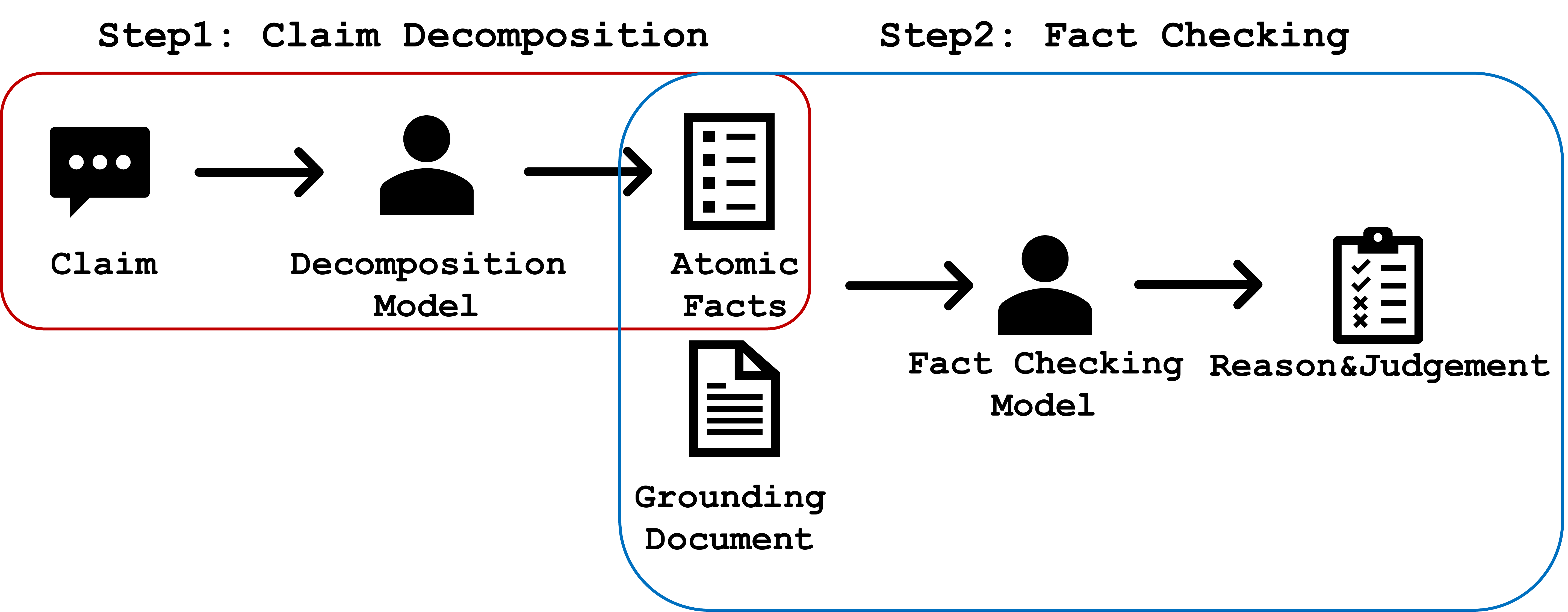}
    \caption{The pipeline for grounded claim checking.}
    \label{fig:workflow}
\end{figure}

We first address a prevalent issue in factuality checking: claims can be complex and aggregate multiple independent pieces of information.
As shown in Figure \ref{fig:workflow}, we decompose the overall evaluation into two simpler steps: 
(1) \textbf{Claim Decomposition}, which breaks a claim into atomic facts, and (2) \textbf{Atomic Fact Checking}, which verifies each fact against the grounding document. 
We describe these steps in detail in the following sections.

\subsubsection{Claim Decomposition}
Claims may contain information from multiple parts of a source document, where supporting evidence can be sparsely distributed.
This increases the difficulty of factuality evaluation.
To simplify the task, we prompt an LLM to decompose each claim into a set of atomic facts that can be verified individually.
To do this, we adopt a few-shot prompt from \citet{tang-etal-2024-minicheck} to guide claim decomposition. 
The prompt is shown in Table \ref{tab:de_prompt}.

\subsubsection{Fact Checking}
During the language learning process, True/False reading comprehension practice is often applied to assess whether students correctly understand a given article. 
Instead of inferring an answer in an unconstrained manner, human examinees may use structured strategies to improve both efficiency and accuracy \citep{yapp-2023-effects,al2021empirical}. 
In practice, they often first verify the information that is explicitly mentioned in the grounding document, then verify whether the remaining information can be inferred from it. 
Inspired by this, we design a fact checking prompt that explicitly guides the model’s reasoning process using the following criteria to assess fact $F$ against document $D$:

\begin{enumerate}
    \item \textbf{C1} The object and subject of the claim are mentioned in $D$.
    \item \textbf{C2} The descriptions of the object and the subject are explicitly supported by $D$.
    \item \textbf{C3} The relation between the object and the subject is explicitly supported by $D$.
    \item \textbf{C4} Any information that was not yet verified can be inferred from $D$.
\end{enumerate}

\begin{figure}
    \centering
    \includegraphics[width=0.9\linewidth]{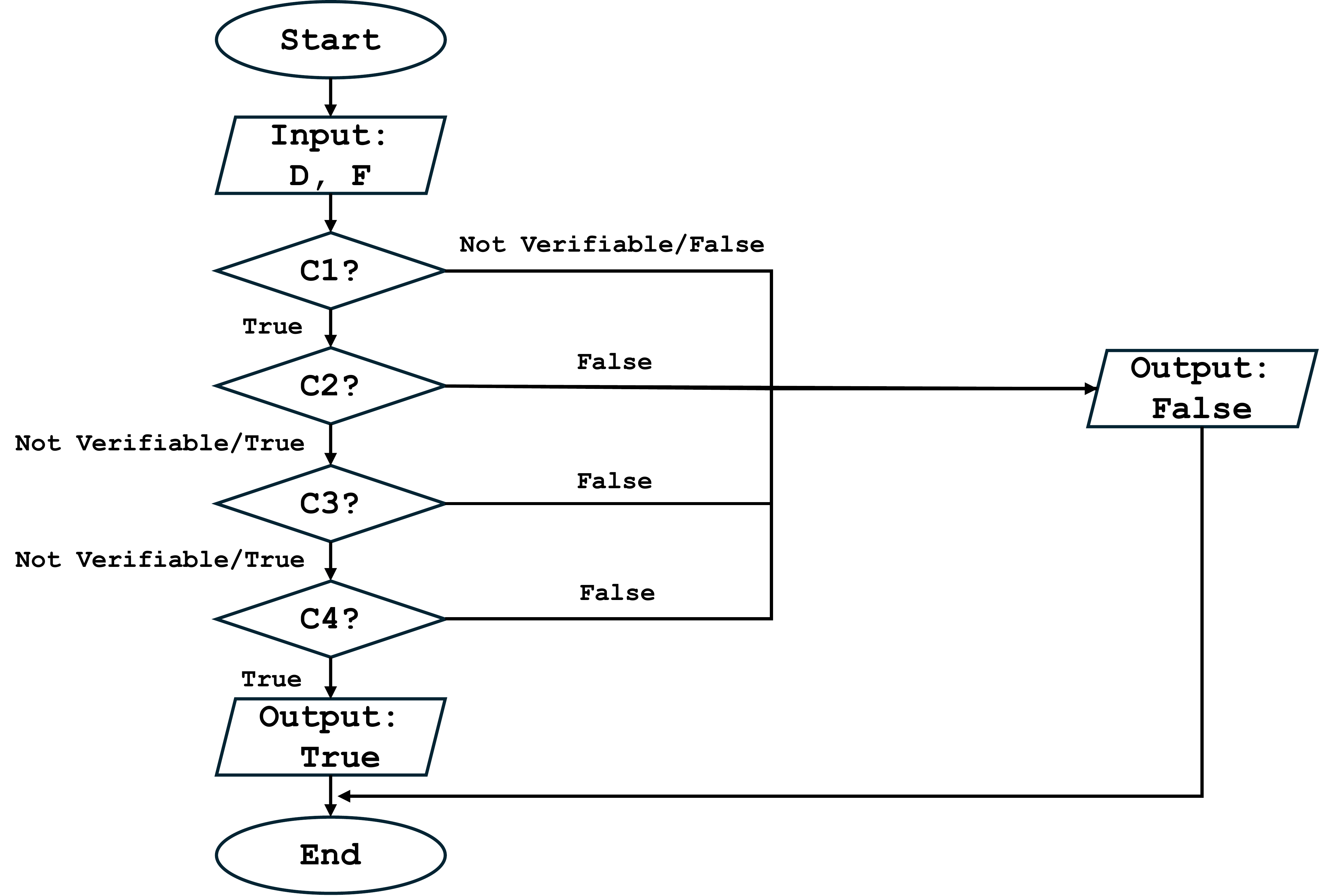}
    \caption{The process for checking a fact, $F$, against each criterion, with grounding document $D$.} \label{fig:criteria}
\end{figure}

Unlike previous work \citep{xu-etal-2024-identifying} that checks different error types individually, our criteria are applied sequentially to form a structured inference process.
It verifies the explicitly mentioned information first, before reasoning whether the as-yet-unverified information is implied by the grounding document, as demonstrated in Figure \ref{fig:criteria}.

For example, given a doc \textit{"Ice can change into liquid water, and water can change into vapour, and vice versa."} and a claim \textit{"Vapour can change into ice."}, both the subject \textit{"vapour"} and the object \textit{"ice"} are mentioned, and there are no descriptions of them to verify, therefore it passes \textbf{C1} and \textbf{C2}. 
For \textbf{C3}, the relation \textit{"can change into ice"} is not explicitly mentioned with \textit{"vapour"}.
Previous methods would recognise it as an erroneous predicate but here we defer it to the next criterion.
The last criterion, \textbf{C4}, analyses the information in two hops. 
\textit{"Vice versa"} implies that the reversed processes are valid, and by combining two reversed processes the claim will be verified as \textit{True} in the end.

In addition to outputting a binary decision (\textit{True/False}), we require the model to provide a rationale. 
This not only has the potential to help users locate evidence in the grounding document, 
but also to improve the consistency of model judgements with the inputs $F$ and $D$.
The full prompt used for this step is shown in Table \ref{tab:re_prompt}.

\subsection{Adapting SLMs for Claim Evaluation}
\label{sec:SLM}
SLMs are less computationally intensive than LLMs, but they are also less capable under the same prompt due to their smaller model scale.
Previous work \citep{qwen3, deepseek-r1} has shown that distillation can substantially enhance SLMs, allowing them to achieve performance comparable to LLMs on certain tasks, including those requiring reasoning abilities.
Motivated by these findings, we train the SLM separately for the two steps to replace the LLM in our evaluation pipeline.

\subsubsection{Claim Decomposition Learning}
Given a claim $c$, we train the SLM to generate decomposed facts that imitate those generated by the teacher LLM.
To construct a large decomposition dataset $D_{De}=\{(c,\{f_{ref}\})\}$, we randomly sample sentences from grounding documents and existing claims, treating each randomly sampled sentence as $c$. 
We then use the teacher model to generate reference facts $\{f_{ref}\}$.

The learning objective can be written as:
\begin{equation*}
    \mathcal{L}(\theta) = \mathbb{E}_{(c, \{f_{ref}\}) \sim D_{De}} \left[ \log P_\theta(\{f_{ref}\} \mid c) \right],
\end{equation*}
where $\theta$ denotes the parameters of the SLM, and $\{f_{ref}\}$ here is considered as a whole string.
We optimise the objective using SFT, aligning the SLM’s generated facts $\{f_{gen}\}$ with the references $\{f_{ref}\}$.

\subsubsection{Fact Checking Learning}
The fact checking step requires the model to generate open-ended rationales, which is difficult to learn from SFT alone, hence we propose a two-stage training approach to better align the SLM with this task. 

\paragraph{Reasoning Format Alignment} Given a grounding document $doc$ and a fact $f$, this training stage focuses on enabling the SLM to generate a rationale $r_{sft}$ that adheres to the test-taking strategies in the prompt.
To create a reasoning dataset, we prompt an LLM to alter $f$ so that it violates certain criteria and use the teacher model to generate the reference rationale $r_{ref}$.
In this way, we create $D_{Re\_SFT}=\{(x,r_{ref})\}$, where $x$ is the prompt that contains $doc$ and $f$.

We apply SFT with the following objective to align the SLM's rationales $r_{sft}$ towards the reference rationales $r_{ref}$:
\begin{equation*}
    \mathcal{L}(\theta) = \mathbb{E}_{D_{Re\_SFT}} \left[ \log P_\theta(r_{ref} \mid x) \right].
\end{equation*}

\paragraph{Mistake Revision} Analogous to human learning, students improve accuracy by practising and correcting mistakes even after learning the key steps. 
Similarly, the SLM from the first stage may still fail to correctly reason about certain facts due to its smaller scale. To address this, we filter the training data to retain only examples where the SLM produces incorrect judgements while the LLM generates correct ones.
This yields a preference dataset $D_{Re\_DPO}=\{(x,y_c,y_r)\}$, where the SLM’s output serves as the rejected completion $y_r$ and the LLM’s output as the chosen completion $y_c$.

We then apply DPO \citep{rafailov2023direct} to align the SLM with the teacher model:
\begin{equation*}
\begin{aligned}
   & \mathcal{L}(\theta) = &\\
   &-\mathbb{E}_{D_{Re\_DPO}} \log \sigma \left[ \beta (s_\theta(x,y_c) - s_\theta(x,y_r)) \right],
\end{aligned}
\end{equation*}
where $\sigma$ is the sigmoid function, $s$ is the log probability that the SLM assigns to the completion, and $\beta$ is a hyperparameter.
This learning process allows the SLM to iteratively revise its mistakes.
Compared to SFT, DPO creates a margin to distinguish rejected and chosen completions, effectively suppressing the mistakes made by the SLM to improve factuality reasoning performance.

\section{Experiments}
Here we evaluate both LLM-based and SLM-based pipelines, including our proposed training method.

\subsection{Benchmarks} 
We conduct evaluation on two factuality benchmarks, FacTax-Benchmark \citep{xu-etal-2024-identifying}\footnote{To avoid confusion with the FacTax metric proposed in the same paper, we refer to it as the FacTax-Benchmark.} and LLM-AggreFact \citep{tang-etal-2024-minicheck}.
FacTax-Benchmark focuses on summarisation factuality checking on news and dialogue data, while LLM-AggreFact includes more diverse source types and claims generated by more recent models including LLMs.
The statistics and breakdowns of the data types of the two benchmarks are listed in Appendix \ref{app:stat}.
Despite the difference, these two benchmarks share overlapping source documents on news summarisation, while differing in the summarisers used to generate the claims.
For FacTax-Benchmark, we exclude the datasets on government reports and stories due to the unavailability of the specific document versions used in the original study. 
As a result, evaluation on this benchmark is conducted only on news \citep{tang-etal-2023-understanding} and dialogue summarisation \citep{zhu-etal-2023-annotating}. 

\subsection{Baselines} 
We compare our method against a range of recent metrics, including TrueTeacher \citep{gekhman-etal-2023-trueteacher}, MiniCheck \citep{tang-etal-2024-minicheck}, FactCG \citep{lei-etal-2025-factcg}, ChatGPT-ZS, ChatGPT-CoT \citep{luo-2023-chatgpt}, FacTax \citep{xu-etal-2024-identifying}, and other LLM-based baselines reported in \citet{tang-etal-2024-minicheck}.\footnote{https://llm-aggrefact.github.io/}
Where available, we cite results from the original papers, which most accurately reflect the performance of corresponding metrics and ensure that we are comparing with the strongest available baselines rather than implementations with less optimal hyperparameters.
Otherwise, we run the released code following their recommended settings. 
For metrics that output continuous scores between 0 and 1 (MiniCheck and FactCG), we follow \citet{tang-etal-2024-minicheck} and apply a fixed threshold of 0.5 to obtain binary True/False predictions (see Appendix \ref{app:threshold_tuning} for results with threshold tuning).
Baseline details are listed in Appendix \ref{app:baseline}.

\subsection{Evaluation Metrics} 

\paragraph{Grounded Claim Factuality Checking} We apply balanced accuracy (BAcc) to evaluate our pipelines, providing a fair comparison to prior works \citep{tang-etal-2024-minicheck,xu-etal-2024-identifying}.
\begin{equation*}
    \text{BAcc}=\frac{1}{2}\left(\frac{TP}{TP+FN}+\frac{TN}{TN+FP}\right),
\end{equation*}
where $TP,TN,FP,FN$ denote the number of true positives, true negatives, false positives, and false negatives respectively.

\paragraph{Claim Decomposition} We evaluate the outcome of training an SLM for the claim decomposition step using ROUGE-L \citep{lin-2004-rouge} and sentence embedding-based cosine similarity (SECS).
\begin{equation*}
    \text{ROUGE-L} = \frac{LCS(F_{gen}, F_{ref})}{|F_{ref}|},
\end{equation*}
\begin{equation*}
    \text{SECS} = \frac{1}{|F_{gen}|}\sum_{F \in F_{gen}}\max_{F' \in F_{ref}} cossim(F,F'),
\end{equation*}
where $F_{gen}, F_{ref}$ refer to the generated and reference facts.
ROUGE-L measures the completeness of the generated facts relative to the reference, while SECS quantifies their semantic similarity.

\begin{table*}[ht]
    \renewcommand{\arraystretch}{1.1}
    \centering
    \resizebox{1.91\columnwidth}{!}{
    \begin{tabular}{|c|l|c|ccccccccc|cc|}
        \hline
        & \multirow{2}{*}{\textbf{Method}} & \multirow{2}{*}{\textbf{Size}} & \multicolumn{4}{c}{\textbf{CNNDM}} & \multicolumn{4}{c}{\textbf{XSUM}} & \textbf{Dialogues} & \multicolumn{2}{c|}{\textbf{Avg}}\\
        \cline{4-14}
        & & &\textbf{Polytope} & \textbf{SummEval} & \textbf{Frank} & \textbf{CLIFF} & \textbf{Wang20} & \textbf{CLIFF} & \textbf{Goyal21} & \textbf{Cao22} & \textbf{DiaSummFact} & \textbf{BAcc ($\uparrow$)} & \textbf{Ranking ($\downarrow$)} \\
        \hline
        \multirow{10}{*}{\rotatebox{90}{Baseline}} & ChatGPT-ZS & / & 90.2 & 79.8 & 54.8 & 65.1 & 71.8 & 74.0 & 63.4 & 68.8 & 66.9 & 70.5 & 9.7\\
        & TrueTeacher & 11B & \cellcolor{pink}100 & 64.9 & 56.1 & 66.6 & \cellcolor{pink}76.6 & 78.7 & 73.6 & 73.3 & 67.4 & 73.0 & 6.5\\
        & FactCG & 0.4B & 48.5 & 64.6 & 56.9 & 76.8 & 67.8 & \cellcolor{pink}81.4 & 68.0 & 70.1 & 68.8 & 67.0 & 9.0\\
        & MiniCheck-DeBERTa & 0.4B & 42.5 & 59.6 & 62.5 & 59.4 & 65.9 & 71.8 & 62.9 & 71.9 & 62.3 & 62.1 & 12.8\\
        & MiniCheck-RoBERTa & 0.4B & 48.5 & 56.6 & 54.5 & 58.9 & 67.4 & 77.2 & 60.8 & 67.7 & 64.1 & 61.7 & 13.5\\
        & MiniCheck-FT5 & 0.8B & 46.3 & 73.7 & 60.6 & 71.1 & 65.9 & 77.3 & 70.0 & 71.0 & 65.8 & 66.9 & 9.6\\
        & MiniCheck-BeSpoke-7B & 7B & 47.8 & 77.6 & 63.4 & 73.2 & \cellcolor{green!50}75.8 & \cellcolor{green!50}80.4 & 73.1 & \cellcolor{pink}74.4 & \cellcolor{green!50}77.1 & 71.4 & \cellcolor{green!50}4.9\\
        & FACTAX-ChatGPT-3.5 & / & 78.4 & 67.4 & 62.8 & 68.7 & 74.1 & 70.3 & \cellcolor{pink}75.1 & 71.7 & 62.8 & 70.1 & 8.5\\
        & FACTAX-ChatGPT-3.5-WD & / & 85.3 & 73.0 & \cellcolor{green!50}67.1 & 70.7 & 71.5 & 70.8 & 68.9 & 69.5 & 64.2 & 71.2 & 8.3\\
        & FACTAX-GPT-4o & / & 94.1 & 79.5 & 61.7 & 79.7 & 74.1 & 70.2 & \cellcolor{green!50}73.9 & 70.1 & 74.6 & \cellcolor{green!50}75.3 & 5.4\\
        \hline
        \multirow{7}{*}{\rotatebox{90}{Ours}}& Qwen3-4B-Thinking & 4B & 87.9 & 70.6 & 53.7 & 67.3 & 72.4 & 70.0 & 57.6 & 65.3 & 76.4 & 69.0 & 10.9\\
        & Qwen3-4B-Instruct & 4B & 73.5 & 87.1 & \cellcolor{pink}67.6 & \cellcolor{green!50}81.3 & 73.9 & 68.2 & 67.3 & 63.0 & 75.0 & 73.0 & 7.1\\
        & Qwen3-30B-A3B-Thinking & 30B & \cellcolor{green!50}95.5 & 74.5 & 63.5 & 75.3 & 67.8 & 72.2 & 66.0 & 67.1 & \cellcolor{pink}78.6 & 73.4 & 6.6\\
        & Qwen3-30B-A3B-Instruct & 30B & 94.1 & \cellcolor{pink}94.4 & 63.7 & \cellcolor{pink}81.4 & 75.7 & 75.4 & 66.9 & \cellcolor{green!50}74.1 & 76.0 & \cellcolor{pink}78.0 & \cellcolor{pink}3.6\\
        \cline{2-14}
        & Qwen3-0.6B & 0.6B & 41.2 & 68.8 & 52.4 & 53.7 & 60.5 & 60.7 & 54.6 & 57.3 & 60.1 & 56.6 & 16.5\\
        & +SFT & 0.6B & 68.8 & 69.3 & 60.6 & 68.9 & 66.5 & 67.9 & 64.9 & 64.2 & 73.7 & 64.8 & 12.1\\
        & +SFT+DPO & 0.6B & 73.7 & \cellcolor{green!50}88.3 & 61.3 & 72.6 & 69.7 & 76.0 & 67.0 & 68.9 & 76.2 & 72.6 & 7.2\\
        \hline
    \end{tabular}
    }
    \caption{Results (BAcc) on FacTax-Benchmark test splits. The \colorbox{pink}{highest} and \colorbox{green!50}{second highest} scores are coloured.}
    \label{tab:factax}
\end{table*}

\begin{table*}[ht]
    \renewcommand{\arraystretch}{1.1}
    \centering
    \resizebox{1.91\columnwidth}{!}{
    \begin{tabular}{|c|l|c|ccccccccccc|cc|}
        \hline
        & \multirow{2}{*}{\textbf{Method}} & \multirow{2}{*}{\textbf{Size}} & \multicolumn{2}{c}{\textbf{AggreFact}} & \multicolumn{2}{c}{\textbf{TOFUEVAL}} & \multirow{2}{*}{\textbf{WICE}} & \multirow{2}{*}{\textbf{Reveal}} & \multirow{2}{*}{\makecell[c]{\textbf{Claim}\\\textbf{Verify}}} & \multirow{2}{*}{\makecell[c]{\textbf{Fact}\\\textbf{Check}}} & \multirow{2}{*}{\makecell[c]{\textbf{Expert}\\\textbf{QA}}} & \multirow{2}{*}{\textbf{LFQA}} & \multirow{2}{*}{\textbf{RAGTruth}} & \multicolumn{2}{c|}{\textbf{Avg}} \\
        \cline{4-7}\cline{15-16}
        & & & \textbf{CNN} & \textbf{XSUM} & \textbf{MediaS} & \textbf{MeetB} &  &  &  &  &  &  &  & \textbf{BAcc ($\uparrow$)} & \textbf{Ranking ($\downarrow$)} \\
        \hline
        \multirow{9}{*}{\rotatebox{90}{Baseline}}& ChatGPT-ZS & / & 63.2 & 72.4 & 66.8 & 73.4 & 68.5 & 84.7 & 65.2 & 70.8 & 57.2 & 73.8 & 75.6 & 70.1 & 13.8\\
        & TrueTeacher & 11B & 60.4 & 74.2 & 70.9 & 73.6 & 64.2 & \cellcolor{pink}91.1 & 64.4 & 76.8 & 59.5 & \cellcolor{pink}90.2 & 80.9 & 73.3 & 8.4 \\
        & FactCG & 0.4B & 70.1 & 73.9 & 72.3 & 74.3 & 74.2 & 88.4 & \cellcolor{pink}78.5 & 72.1 & 59.1 & \cellcolor{green!50}86.7 & 82.3 & 75.6 & 5.8 \\
        & MiniCheck-DeBERTa & 0.4B & 64.2 & 71.0 & 69.3 & 72.7 & 69.4 & 87.3 & 75.6 & 73.0 & 58.9 & 83.9 & 78.8 & 73.1 & 10 \\
        & MiniCheck-RoBERTa & 0.4B & 63.7 & 70.8 & 71.9 & 75.9 & 67.6 & \cellcolor{green!50}88.8 & \cellcolor{green!50}77.4 & 73.3 & 57.4 & 84.4 & 77.2 & 73.5 & 9.1\\
        & MiniCheck-FT5 & 0.8B & 69.9 & 74.3 & 73.6 & 77.3 & 72.2 & 86.2 & 74.6 & 74.7 & 59.0 & 85.2 & 78.0 & 75.0 & 6.8\\
        & MiniCheck-BeSpoke-7B & 7B & 65.5 & \cellcolor{pink}77.8 & \cellcolor{pink}76.0 & 78.3 & \cellcolor{pink}83.0 & 88.0 & 75.3 & 77.7 & 59.2 & \cellcolor{green!50}86.7 & \cellcolor{green!50}84.0 & \cellcolor{pink}77.4 & \cellcolor{pink}3.3\\
        & Llama-3.3-70B-Instruct & 70B & 68.7 & 74.7 & 69.5 & 78.4 & 76.6 & 85.5 & 67.4 & \cellcolor{green!50}78.5 & 58.3 & 79.8 & 82.6 & 74.5 & 7.3\\
        & GPT-4o-2024-05-13 & / & 68.1 & \cellcolor{green!50}76.8 & 71.4 & \cellcolor{green!50}79.8 & \cellcolor{green!50}78.5 & 86.5 & 69.0 & 77.5 & \cellcolor{green!50}59.6 & 83.6 & \cellcolor{pink}84.3 & 75.9 & 4.8\\
        \hline
        \multirow{7}{*}{\rotatebox{90}{Ours}} & Qwen3-4B-Thinking & 4B & 68.0 & 69.8 & 69.8 & 75.3 & 73.1 & 84.6 & 71.5 & 74.0 & 57.3 & 77.8 & 81.0 & 73.1 & 11\\
        & Qwen3-4B-Instruct & 4B & 70.8 & 70.0 & \cellcolor{green!50}73.7 & 79.3 & 75.4 & 88.3 & 72.9 & 75.3 & 59.1 & 82.4 & \cellcolor{green!50}84.0 & 75.6 & 5.3\\
        & Qwen3-30B-A3B-Thinking & 30B & \cellcolor{green!50}71.7 & 72.9 & 73.6 & 78.2 & 69.8 & 85.7 & 73.2 & 72.7 & 58.1 & 81.2 & 82.1 & 74.5 & 8\\
        & Qwen3-30B-A3B-Instruct & 30B & \cellcolor{pink}72.5 & 74.4 & 70.6 & \cellcolor{pink}81.8 & 77.0 & 88.4 & 72.7 & \cellcolor{pink}79.4 & \cellcolor{pink}60.6 & 79.0 & 82.6 & \cellcolor{green!50}76.3 & \cellcolor{green!50}4\\
        \cline{2-16}
        & Qwen3-0.6B-Instruct & 0.6B & 51.5 & 51.6 & 55.2 & 55.4 & 52.1 & 70.4 & 54.0 & 61.3 & 52.3 & 58.2 & 57.6 & 56.3 & 16\\
        & +SFT & 0.6B & 65.1 & 71.3 & 66.2 & 72.4 & 73.0 & 86.4 & 66.2 & 73.6 & 58.7 & 75.7 & 75.9 & 71.3 & 12\\
        & +SFT+DPO & 0.6B & 69.8 & 72.2 & 68.7 & 75.8 & 75.4 & 86.1 & 71.3 & 75.2 & 58.7 & 77.9 & 77.1 & 73.6 & 9.5\\
        \hline
    \end{tabular}
    }
    \caption{Results (BAcc) on LLM-AggreFact test splits. The \colorbox{pink}{highest} and \colorbox{green!50}{second highest} scores are coloured.}
    \label{tab:llm-aggrefact}
\end{table*}

\subsection{Setup and Implementations}
\paragraph{LLM-based Pipeline} We apply our prompt to a series of Qwen3 models \citep{qwen3} because of their outstanding instruction-following ability and long context window (detailed analysis in Appendix \ref{app:long_context}). 
Specifically, 
we use Qwen3-4B-Instruct-2507\footnote{https://huggingface.co/Qwen/Qwen3-4B-Instruct-2507} and Qwen3-30B-A3B-Instruct-2507\footnote{https://huggingface.co/Qwen/Qwen3-30B-A3B-Instruct-2507} in the LLM-based pipeline.
We also report results from the \textit{Thinking} variants, although our prompt does not require long chain-of-thought reasoning.

\paragraph{SLM Fine-tuning Implementations} We apply Qwen3-0.6B\footnote{https://huggingface.co/Qwen/Qwen3-0.6B} in the SLM-based pipeline.
During SFT, we use a learning rate of $1\times10^{-4}$ to adapt the output format. 
By the DPO stage, the model has learned natural language reasoning and criterion-checking abilities through SFT. 
Therefore, we apply a smaller learning rate of $1\times10^{-7}$ to revise the chain-of-thought content while preserving the learned behaviour. 
All models are trained for $3$ epochs with early stopping to mitigate overfitting. 
Previous work \citep{tang-etal-2024-minicheck} used the development split to curate the training data.
We follow it in our work and use the two development splits to create $D_{De}$ and $D_{Re\_SFT}$.
Their statistics are reported in Tables \ref{tab:de_stat} and \ref{tab:re_stat}. 
All prompts used are provided in Appendix \ref{app:prompt}.

\subsection{Results}
We first present the performance of all methods on the two benchmarks, then assess the training outcome of SLMs.

\subsubsection{Claim Checking Performance}
\paragraph{LLM-based Pipeline} Results in Tables \ref{tab:factax} and \ref{tab:llm-aggrefact} demonstrate that our prompt successfully adapts a general purpose LLM into an off-the-shelf factuality evaluation metric in a zero-shot setting, avoiding the need for training data. 
On FacTax-Benchmark, our Qwen3-30B-A3B-Instruct pipeline achieves state-of-the-art in both average score and overall ranking.
On LLM-AggreFact, our method ranks second in Table \ref{tab:llm-aggrefact} to a metric that requires post-training an LLM, while scoring highest on four subsets.
Consistent performance across benchmarks highlights the effectiveness and robustness of our prompt, without requiring any fine-tuning.
Notably, our method outperforms methods based on closed-source models or larger open-source models such as GPT-4o and Llama-3.3-70B. 

\paragraph{SLM-based Pipeline} The bottom three rows in both tables illustrate the contribution of each training stage.
The SLM consistently benefits from SFT and DPO, outperforming ChatGPT-3.5–based baselines on both benchmarks and achieving performance comparable to TrueTeacher, which relies on a substantially larger 11B base model and a 540B teacher model.
Compared to metrics using small backbones (<1B parameters), our SLM-based pipeline surpasses all on FacTax-Benchmark, while showing a small gap to stronger baselines such as FactCG and MiniCheck-FT5 on LLM-AggreFact.
This gap primarily occurs on datasets with complex grounding documents, including LFQA and TOFUEVAL-MediaS, where they challenge generative models’ reasoning and comprehension. 
Notably, our teacher model also exhibits degraded performance on these subsets, suggesting a potential direction for future research.

\paragraph{Impact of Backbone Size} For metrics trained on the same data or using backbone models within a model family, performance consistently improves with increasing model size. 
For example, our prompt achieves higher scores with the 30B backbone than with the 4B backbone in both thinking and instruct modes, and similar trends are observed for MiniCheck. 
However, model size alone does not determine performance. 
Our 4B model outperforms Llama-3.3-70B-Instruct when the latter is used with a direct prompt on LLM-AggreFact, suggesting the importance of prompt design for zero-shot LLM-based factuality metrics.

\subsubsection{SLM Claim Decomposition Quality}
Table \ref{tab:decomposition} shows that the SLM achieves substantial improvements after claim decomposition training.
We further study the effect of the backbone model in this step.
Table \ref{tab:cd_quality} compares the DPO checkpoint’s balanced accuracy using facts decomposed by the 30B LLM versus the SFT-trained checkpoint.
The small performance gap indicates that the trained SLM checkpoint can reliably decompose a claim into atomic facts, providing a strong foundation for training the subsequent atomic fact checker.
\begin{table}[!htbp]
    \centering
    \resizebox{0.9\columnwidth}{!}{
    \begin{tabular}{ccccc}
    \toprule
         & \multicolumn{2}{c}{\textbf{FacTax-Benchmark}} & \multicolumn{2}{c}{\textbf{LLM-AggreFact}} \\
    \cmidrule{2-5}
         & \textbf{ROUGE-L ($\uparrow$)} & \textbf{SECS ($\uparrow$)} & \textbf{ROUGE-L ($\uparrow$)} & \textbf{SECS ($\uparrow$)} \\
    \midrule
     Before & 63.3 & 71.5 & 54.0 & 58.7\\
     After & 89.3 & 95.5 & 84.3 & 92.6\\
    \bottomrule
    \end{tabular}
    }
    \caption{The SLM evaluation results before and after training for the claim decomposition task.}
    \label{tab:decomposition}
\end{table}
\begin{table}[!htbp]
    \footnotesize
    \centering
    \resizebox{0.9\columnwidth}{!}{
    \begin{tabular}{ccc}
    \toprule
     \makecell[c]{\textbf{Claim}\\\textbf{Decomposition}}& \textbf{FacTax-Benchmark} & \textbf{LLM-AggreFact} \\
    \midrule
      30B-A3B & 73.4 & 73.7\\
      0.6B & 72.6 & 73.6\\
    \bottomrule
    \end{tabular}
    }
    \caption{Mean BAcc of the SLM pipeline with different claim decomposition models}
    \label{tab:cd_quality}
\end{table}

\section{Analysis}

\subsection{Impact of Output Length} 
We find that switching the thinking mode on does not benefit the performance of our LLM-based pipeline.
Moreover, Table \ref{tab:tokens} shows that instruct mode reduces token usage by roughly 90\% on FacTax-Benchmark and over 80\% on LLM-AggreFact. 
These results suggest that effective factuality judgements can be achieved through guided reasoning with explicitly defined steps, rather than unrestricted exploration.
It demonstrates that our prompting strategy provides clear and practical guidance for LLMs, leading to stronger performance and lower inference cost.

\begin{table}[ht]
    \centering
    \resizebox{0.9\columnwidth}{!}{
    \begin{tabular}{llcc|cc}
    \toprule
     \multirow{2}{*}{\textbf{Qwen3}} & \multirow{2}{*}{\textbf{Mode}} & \multicolumn{2}{c|}{\textbf{FacTax}} & \multicolumn{2}{c}{\textbf{LLM-AggreFact}} \\
     \cmidrule{3-6}
      &  & \textbf{TokenUsed} & \textbf{Ratio (\%)} & \textbf{TokenUsed} & \textbf{Ratio (\%)} \\
    \midrule
     \multirow{2}{*}{4B} & Instruct  & 2803.6 & \multirow{2}{*}{10.4} & 2132.5 & \multirow{2}{*}{12.5}\\
      & Thinking  & 27001.9 &  & 17025.4\\
    \midrule
     \multirow{2}{*}{30B-A3B} & Instruct & 1152.4 & \multirow{2}{*}{10.5}  & 1024.4 & \multirow{2}{*}{17.7}\\
      & Thinking  & 10924.8 & & 5781.7\\
    \bottomrule
    \end{tabular}
    }
    \caption{The impact of switching thinking on and off. The ratio column presents the percentage of the tokens used in the instruct mode over the thinking mode.}
    \label{tab:tokens}
\end{table}

\subsection{Ablation Study}
\paragraph{LLM-based Pipeline} Table \ref{tab:ablation_zero} reports the results of removing the claim decomposition step and the test-taking strategies from the prompt. 
When both components are removed we prompt the model with the claim and directly ask the model to judge factuality under instruct mode.

Removing either component leads to a noticeable drop in balanced accuracy on both benchmarks.
This indicates that both claim decomposition and guided reasoning via test-taking strategies are critical to the effectiveness of our approach.

\begin{table}[htbp]
    \centering
    \resizebox{0.9\columnwidth}{!}{
    \begin{tabular}{llcc}
    \toprule
     \textbf{Qwen3} & \textbf{Pipeline} & \textbf{FacTax-Benchmark} & \textbf{LLM-AggreFact} \\
    \midrule
     \multirow{4}{*}{4B} & Full & 73.0 ($\pm$ 0.2)& 75.6 ($\pm$ 0.4)\\
      & -Decomposition & 72.3 ($\pm$ 0.4)& 74.6 ($\pm$ 0.3)\\
      & -Strategy & 71.6 ($\pm$ 0.3)& 73.1 ($\pm$ 0.3)\\
      & -Both & 69.4 ($\pm$ 0.5)& 72.1 ($\pm$ 0.2)\\
    \midrule
     \multirow{4}{*}{30B-A3B} & Full & 78.0 ($\pm$ 0.3) & 76.3 ($\pm$ 0.4)\\
      & -Decomposition & 76.2 ($\pm$ 0.4) & 76.0 ($\pm$ 0.4)\\
      & -Strategy & 76.5 ($\pm$ 0.4)& 75.9 ($\pm$ 0.3)\\
      & -Both & 74.8 ($\pm$ 0.3)& 74.7 ($\pm$ 0.4)\\
    \bottomrule
    \end{tabular}
    }
    \caption{Ablation study for the LLM-based pipeline (mean BAcc). Differences between the full model and ablated variants are statistically significant ($p<0.05$) based on a paired bootstrap test. Standard deviations are computed from three random seeds.}
    \label{tab:ablation_zero}
\end{table}

Table \ref{tab:ablation_criterion} below presents the results of a more fine-grained ablation study over criteria. 
Criteria C1–C3 jointly help the model focus on evidence in the document, while C4 enables the model to perform reasonable inference based on the available information. 
Therefore, we compare the performance of the full metric with that of using only C1–C3 and only C4.
To be more specific, we instruct the model not to perform inference when using only C1–C3. 
For the C4-only prompt, we remove C1–C3 and allow the model to perform inference based on the whole grounding document. 
The results show that the evidence-locating criteria (C1–C3) and the inference criterion (C4) are complementary and mutually beneficial. 
When combined, they achieve the best performance on both benchmarks.

\begin{table}[htbp]
    \centering
    \resizebox{0.9\columnwidth}{!}{
    \begin{tabular}{llcc}
    \toprule
     \textbf{Qwen} & \textbf{Criteria} & \textbf{FacTax-Benchmark} & \textbf{LLM-AggreFact} \\
    \midrule
     \multirow{3}{*}{4B} & Full & 73.0 ($\pm$ 0.2) & 75.6 ($\pm$ 0.4)\\
      & C1+C2+C3 & 72.4 ($\pm$ 0.3)& 73.4 ($\pm$ 0.2)\\
      & C4 & 71.9 ($\pm$ 0.3)& 74.5 ($\pm$ 0.5)\\
    \midrule
     \multirow{3}{*}{30B-A3B} & Full & 78.0 ($\pm$ 0.3)& 76.3 ($\pm$ 0.4)\\
      & C1+C2+C3 & 76.0 ($\pm$ 0.4)& 75.9 ($\pm$ 0.4)\\
      & C4 & 76.2 ($\pm$ 0.6)& 73.2 ($\pm$ 0.3)\\
    \bottomrule
    \end{tabular}
    }
    \caption{Ablation study for the criteria in the test-taking strategy (mean BAcc). Differences between the full model and ablated variants are statistically significant ($p<0.05$) based on a paired bootstrap test. Standard deviations are computed from three random seeds.}
    \label{tab:ablation_criterion}
\end{table}

\paragraph{SLM Training} 
Tables \ref{tab:factax} and \ref{tab:llm-aggrefact} show the contribution of each training stage to SLM performance. 
We also investigate the choice of dataset used to create the distillation datasets $D_{De}$ and $D_{Re\_SFT}$ 
and repeat SFT while omitting the training data derived from each of the benchmarks in turn. 
The dataset created from FacTax-Benchmark is noted as $D_{FacTax}$, and that from LLM-AggreFact as $D_{LA}$.

Table \ref{tab:ablation_slm} shows that removing the training data corresponding to the benchmark substantially degrades performance on that benchmark. 
This suggests that the SLM’s generalisation ability is more limited than that of the LLM, highlighting the importance of providing diverse training examples.
\begin{table}[!htbp]
    \footnotesize
    \centering
    \resizebox{0.85\columnwidth}{!}{
    \begin{tabular}{lcc}
    \toprule
     \textbf{Pipeline} & \textbf{FacTax-Benchmark} & \textbf{LLM-AggreFact} \\
    \midrule
     SFT & 67.2 ($\pm$ 0.5)& 71.3 ($\pm$ 0.4)\\
     -$D_{FacTax}$ & 64.2 ($\pm$ 0.3)& 66.7 ($\pm$ 0.7)\\
     -$D_{LA}$ & 65.3 ($\pm$ 0.6)& 62.1 ($\pm$ 0.4)\\
    \bottomrule
    \end{tabular}
    }
    \caption{Ablation study on SLM training data origins (mean BAcc). Differences between the full model and ablated variants are statistically significant ($p<0.05$) based on a paired bootstrap test. Standard deviations are computed from three random seeds.}
    \label{tab:ablation_slm}
\end{table}

\subsection{Comparison with Single-pass Checking}
We study the necessity of splitting the fact-checking process into two stages by comparing its performance against a 
single prompt that incorporates the entire claim decomposition and fact checking stages in a single pass. 
Table \ref{tab:single_pass} shows the benefits of splitting the pipeline into two stages.
The two-step pipeline ensures that the model focuses on the information required at each stage, reducing potential distractions buried in long chain-of-thought outputs from the single-pass pipeline.
The single-pass prompt is reported in Table \ref{tab:single_pass_prompt}.

\begin{table}[htbp]
    \footnotesize
    \centering
    \resizebox{0.95\columnwidth}{!}{
    \begin{tabular}{lccc}
    \toprule
     & \textbf{Size} & \textbf{FacTax-Benchmark} & \textbf{LLM-AggreFact} \\
    \midrule
     \multirow{2}{*}{SinglePass}& 4B & 70.2 & 73.7 \\
      & 30B & 72.3 & 74.4 \\
    \midrule
     \multirow{2}{*}{Ours} & 4B & 73.0 & 75.6 \\
     & 30B & 78.0 & 76.3\\
    \bottomrule
    \end{tabular}
    }
    \caption{
    Comparison against a single-pass prompt.}
    \label{tab:single_pass}
\end{table}

\subsection{Prompt Paraphrase}
The human test-taking strategy applied in this work is a widely used skill developed specifically for solving reading comprehension problems in language tests, particularly among second-language learners \citep{yapp-2023-effects,al2021empirical}.
While the specific descriptions and wording may vary across instructors, the core objective remains the same. 
The final aim is to efficiently identify a claim's evidence in the text and verify the claim against the evidence. 
To demonstrate that our strategy is robust to the variations in wording, 
we ask ChatGPT and Gemini (Fast) to paraphrase the criteria.
The paraphrased prompts are shown in Tables \ref{tab:chatgpt_prompt} and \ref{tab:gemini_prompt}.
Results in Table \ref{tab:prompt_paraphrase} show that the balanced accuracy remains the same level after paraphrasing the prompt, indicating that our strategy is not sensitive to the prompt wording.

\begin{table}[htbp]
    \footnotesize
    \centering
    \resizebox{0.95\columnwidth}{!}{
    \begin{tabular}{lccc}
    \toprule
     & \textbf{Size} & \textbf{FacTax-Benchmark} & \textbf{LLM-AggreFact} \\
    \midrule
     \multirow{2}{*}{ChatGPT}& 4B & 72.3 & 75.7 \\
      & 30B & 79.4 & 75.9 \\
    \midrule
     \multirow{2}{*}{Gemini}& 4B & 73.9 & 75.8 \\
      & 30B & 78.6 & 75.8 \\
    \midrule
     \multirow{2}{*}{Ours} & 4B & 73.0 & 75.6 \\
     & 30B & 78.0 & 76.3\\
    \bottomrule
    \end{tabular}
    }
    \caption{Performance after using ChatGPT and Gemini to paraphrase our prompt.}
    \label{tab:prompt_paraphrase}
\end{table}

\subsection{Rationale Examples} A key advantage of our pipeline is its ability to generate explanations for factuality judgements. 
Unlike metrics that output a single scalar score, these explanations help users identify inconsistent parts of a claim. 
While some LLM-based metrics also produce reasoning, their outputs are often excessively long, thus less interpretable. 
Example outputs in Table \ref{tab:quality} show that the atomic facts generated by the LLM and SLM differ by only a single word, and both fact checking processes follow the prompt well, pinpointing exactly which part of the source text supports the fact.

By examining examples with incorrect judgments, we observe that the model can sometimes be overly strict when making decisions (mostly in C3 and C4). 
In Table \ref{tab:rebuttal_example}, the model correctly recognises that the fact is verifiable but becomes uncertain due to subtle wording differences between the two pieces of text, which leads to an incorrect prediction.
The generated rationale is still reasonable, but overemphasises minor semantic differences. 
This example demonstrates the value of generating rationales that reveal the process leading to the final decision.   

\section{Conclusion}
In this work, we reframe grounded claim factuality checking as a reading comprehension task.
Unlike prior approaches that directly query a model for claim support, we utilise human test-taking strategies in the prompt to guide the chain-of-thoughts. 
Experiments show that this prompt effectively adapted a general purpose LLM into an off-the-shelf factuality evaluation metric, avoiding the need for any training data.
To reduce computational cost, we further distil the LLM into separate SLMs for claim decomposition and fact checking. 
Inspired by how students improve by correcting their mistakes, the SLMs are additionally trained to revise their own errors to enhance reasoning accuracy.
The resulting SLMs achieve performance comparable to much larger LLMs across two benchmarks, demonstrating the effectiveness of our human-inspired training method.
Overall, our study shows that, given suitable training strategies, SLMs can perform claim factuality checking efficiently while maintaining interpretability, though there remains potential for further improvement to address limitations brought by model scale, such as reasoning and generalisation ability.

\section*{Limitations}
\paragraph{Base Model Ability} Although our SLM-based pipeline outperforms several metrics that rely on LLM backbones, further performance improvements may require stronger base models. 
While this work demonstrates an effective approach for training SLMs for claim factuality checking, the resulting models remain constrained by their limited scale, particularly in terms of complex reasoning and deep language understanding.

\paragraph{Teacher Model Scale} We use a 30B model as the teacher to generate reference outputs for distillation, which is not the strongest model available in the Qwen3 family. 
Since the quality of reference data plays a critical role in distillation, performance could potentially be improved by employing a stronger teacher model or by mixing outputs from multiple teacher models to enhance data diversity and quality.

\paragraph{Prompt Engineering} LLMs are known to be sensitive to prompt design. 
A strength of our approach is the use of a zero-shot prompt for fact checking, which avoids reliance on and sensitivity to few-shot examples.
However, performance may still vary with different descriptions of the test-taking strategies in the prompt. 
While it is impractical to exhaustively evaluate all possible prompt wordings, performance could be boosted by further systematic prompt refinement, but this would bring a risk of over-fitting to specific models and datasets, which we have avoided in this paper. Furthermore, our experiments in Table \ref{tab:prompt_paraphrase} show only small variations in performance with different prompt wordings. 

\section*{Acknowledgments}
The authors acknowledge the use of resources provided by the Isambard-AI National AI Research Resource (AIRR). Isambard-AI is operated by the University of Bristol and is funded by the UK Government’s Department for Science, Innovation and Technology (DSIT) via UK Research and Innovation; and the Science and Technology Facilities Council [ST/AIRR/I-A-I/1023]. The financial support for Yuxuan Ye is provided by the programme of the China Scholarship Council (No. 202108060154). Raul Santos-Rodriguez is supported by the UKRI Turing AI Fellowship EP/V024817/1.


\bibliography{custom,anthology-1,anthology-2}

@inproceedings{lin-2004-rouge,title = "{ROUGE}: A Package for Automatic Evaluation of Summaries",author = "Lin, Chin-Yew",booktitle = "Text Summarization Branches Out",month = jul,year = "2004",address = "Barcelona, Spain",publisher = acl,url = anth # {W04-1013/},pages = "74--81"}

@article{rafailov2023direct,
  title={Direct preference optimization: Your language model is secretly a reward model},
  author={Rafailov, Rafael and Sharma, Archit and Mitchell, Eric and Manning, Christopher D and Ermon, Stefano and Finn, Chelsea},
  journal={Advances in neural information processing systems},
  volume={36},
  pages={53728--53741},
  year={2023}
}

@article{al2021empirical,
  title={An Empirical Study on the Effect of Instruction on Metacognitive Strategies on EFL Reading Comprehension: The Case of Foundation-Level Students in Oman.},
  author={Al-Kiyumi, Omaima and Seyabi, Fawzia Al and Hassan, Abdul Hamid},
  journal={International Education Studies},
  volume={14},
  number={8},
  pages={30--42},
  year={2021},
  publisher={ERIC}
}

@misc{qwen3,
      title={Qwen3 Technical Report}, 
      author={QwenTeam},
      year={2025},
      eprint={2505.09388},
      archivePrefix={arXiv},
      primaryClass={cs.CL},
      url={https://arxiv.org/abs/2505.09388}, 
}

@misc{deepseek-r1,
      title={DeepSeek-R1: Incentivizing Reasoning Capability in LLMs via Reinforcement Learning}, 
      author={DeepSeek-AI},
      year={2025},
      eprint={2501.12948},
      archivePrefix={arXiv},
      primaryClass={cs.CL},
      url={https://arxiv.org/abs/2501.12948}, 
}

@article{luo-2023-chatgpt,
  title={Chatgpt as a factual inconsistency evaluator for text summarization},
  author={Luo, Zheheng and Xie, Qianqian and Ananiadou, Sophia},
  journal={arXiv preprint arXiv:2303.15621},
  year={2023}
}

@article{dubey-2024-llama,
  title={The llama 3 herd of models},
  author={Dubey, Abhimanyu and Jauhri, Abhinav and Pandey, Abhinav and Kadian, Abhishek and Al-Dahle, Ahmad and Letman, Aiesha and Mathur, Akhil and Schelten, Alan and Yang, Amy and Fan, Angela and others},
  journal={arXiv e-prints},
  pages={arXiv--2407},
  year={2024}
}

@article{fabbri-2021-summeval,
  title={Summeval: Re-evaluating summarization evaluation},
  author={Fabbri, Alexander R and Kry{\'s}ci{\'n}ski, Wojciech and McCann, Bryan and Xiong, Caiming and Socher, Richard and Radev, Dragomir},
  journal={Transactions of the Association for Computational Linguistics},
  volume={9},
  pages={391--409},
  year={2021},
  publisher={MIT Press One Rogers Street, Cambridge, MA 02142-1209, USA journals-info~…}
}

@article{zhang-2019-bertscore,
  title={Bertscore: Evaluating text generation with bert},
  author={Zhang, Tianyi and Kishore, Varsha and Wu, Felix and Weinberger, Kilian Q and Artzi, Yoav},
  journal={arXiv preprint arXiv:1904.09675},
  year={2019}
}

@article{ye-2024-using,
  title={Using similarity to evaluate factual consistency in summaries},
  author={Ye, Yuxuan and Simpson, Edwin and Rodriguez, Raul Santos},
  journal={arXiv preprint arXiv:2409.15090},
  year={2024}
}

@article{he-2020-deberta,
  title={Deberta: Decoding-enhanced bert with disentangled attention},
  author={He, Pengcheng and Liu, Xiaodong and Gao, Jianfeng and Chen, Weizhu},
  journal={arXiv preprint arXiv:2006.03654},
  year={2020}
}

@article{achiam-2023-gpt,
  title={Gpt-4 technical report},
  author={Achiam, Josh and Adler, Steven and Agarwal, Sandhini and Ahmad, Lama and Akkaya, Ilge and Aleman, Florencia Leoni and Almeida, Diogo and Altenschmidt, Janko and Altman, Sam and Anadkat, Shyamal and others},
  journal={arXiv preprint arXiv:2303.08774},
  year={2023}
}

@InProceedings{zhang-2024-how,
  title = 	 {How Language Model Hallucinations Can Snowball},
  author =       {Zhang, Muru and Press, Ofir and Merrill, William and Liu, Alisa and Smith, Noah A.},
  booktitle = 	 {Proceedings of the 41st International Conference on Machine Learning},
  pages = 	 {59670--59684},
  year = 	 {2024},
  editor = 	 {Salakhutdinov, Ruslan and Kolter, Zico and Heller, Katherine and Weller, Adrian and Oliver, Nuria and Scarlett, Jonathan and Berkenkamp, Felix},
  volume = 	 {235},
  series = 	 {Proceedings of Machine Learning Research},
  month = 	 {21--27 Jul},
  publisher =    {PMLR},
  url = 	 {https://proceedings.mlr.press/v235/zhang24ay.html}
}

@article{yapp-2023-effects,
  title={Effects of reading strategy instruction in English as a second language on students’ academic reading comprehension},
  author={Yapp, Deborah and de Graaff, Rick and van den Bergh, Huub},
  journal={Language Teaching Research},
  volume={27},
  number={6},
  pages={1456--1479},
  year={2023},
  publisher={SAGE Publications Sage UK: London, England}
}

@article{chen-2023-understanding,
  title={Understanding retrieval augmentation for long-form question answering},
  author={Chen, Hung-Ting and Xu, Fangyuan and Arora, Shane and Choi, Eunsol},
  journal={arXiv preprint arXiv:2310.12150},
  year={2023}
}

@article{hsieh-2024-ruler,
  title={RULER: What's the Real Context Size of Your Long-Context Language Models?},
  author={Cheng-Ping Hsieh and Simeng Sun and Samuel Kriman and Shantanu Acharya and Dima Rekesh and Fei Jia and Yang Zhang and Boris Ginsburg},
  year={2024},
  journal={arXiv preprint arXiv:2404.06654},
}

\appendix

\section{Baseline Details}
\label{app:baseline}
We introduce the details of all the mentioned baselines in this paper below.

\paragraph{TrueTeacher} distils factuality judgement ability from a directly prompted 540B LLM into a relatively smaller model with 11B parameters. While effective, both the teacher and student models are substantially larger than our SLM.
\paragraph{MiniCheck} synthesises training data conditioned on either the grounding document or the claim, which leads to substantially improved entailment classifiers. It is a strong baseline that achieves performance comparable to much larger LLMs.
\paragraph{FactCG} is specifically trained to improve reasoning over claims that require multi-hop inference. It represents the current state of the art among entailment-based metrics and is built on BERT-style architectures such as DeBERTa \citep{he-2020-deberta}.
\paragraph{ChatGPT-\{ZS,CoT\}} directly prompt ChatGPT-3.5-turbo to judge whether a summary is consistent with the source document. The chain-of-thought (CoT) variant additionally encourages step-by-step reasoning, resulting in an open-ended reasoning process. Other LLM-based baselines, if mentioned directly with their names in this paper, follow a similar method but differ in base models.
\paragraph{FACTAX} defines fine-grained error categories in the prompt and uses LLMs’ reasoning capabilities to decide if the claim contains certain error types.

\section{Long-context Handling}
\label{app:long_context}
We use Qwen3 \citep{qwen3} in this work, which not only simply supports long context windows but has also demonstrated strong long-context handling ability in prior evaluations. 
For example, the 30B-A3B variant outperforms strong close-source models such as GPT-4 and Gemini-Pro-1.5 on RULER \citep{hsieh-2024-ruler}, a benchmark designed to evaluate a model’s ability to locate required information within extended contexts (needle-in-a-haystack task), which aligns with our application for locating supporting evidence in this paper. 
Importantly, the average source document lengths in our benchmarks are substantially shorter than the evaluated context ranges (e.g., 395.3 tokens for FacTax and 559.4 tokens for LLM-AggreFact, with similar values on the development splits), and therefore fall well within the range where long-context degradation effects are expected to be limited. 

We also note that many earlier BERT-based factuality metrics rely on chunking partly because they are adapted from natural language inference tasks, where inputs often consist of sentence-level (premise, hypothesis) pairs. 
When applied to document-level grounding, such models may exhibit unstable behaviour without segmentation. 
In contrast, generative LLM-based metrics operate natively on longer sequences, which reduces the need for aggressive chunking.

Our experimental results on both benchmarks demonstrate that the proposed no-chunking setup is practically feasible. 
The general-purpose LLM-based zero-shot fact-checker achieves competitive performance and is only outperformed by models that are specifically trained for safety or alignment judgment. 
These findings suggest that, at least within the document length ranges considered, eliminating chunking does not degrade evaluation quality.

\begin{figure*}
    \centering
    \includegraphics[width=\linewidth]{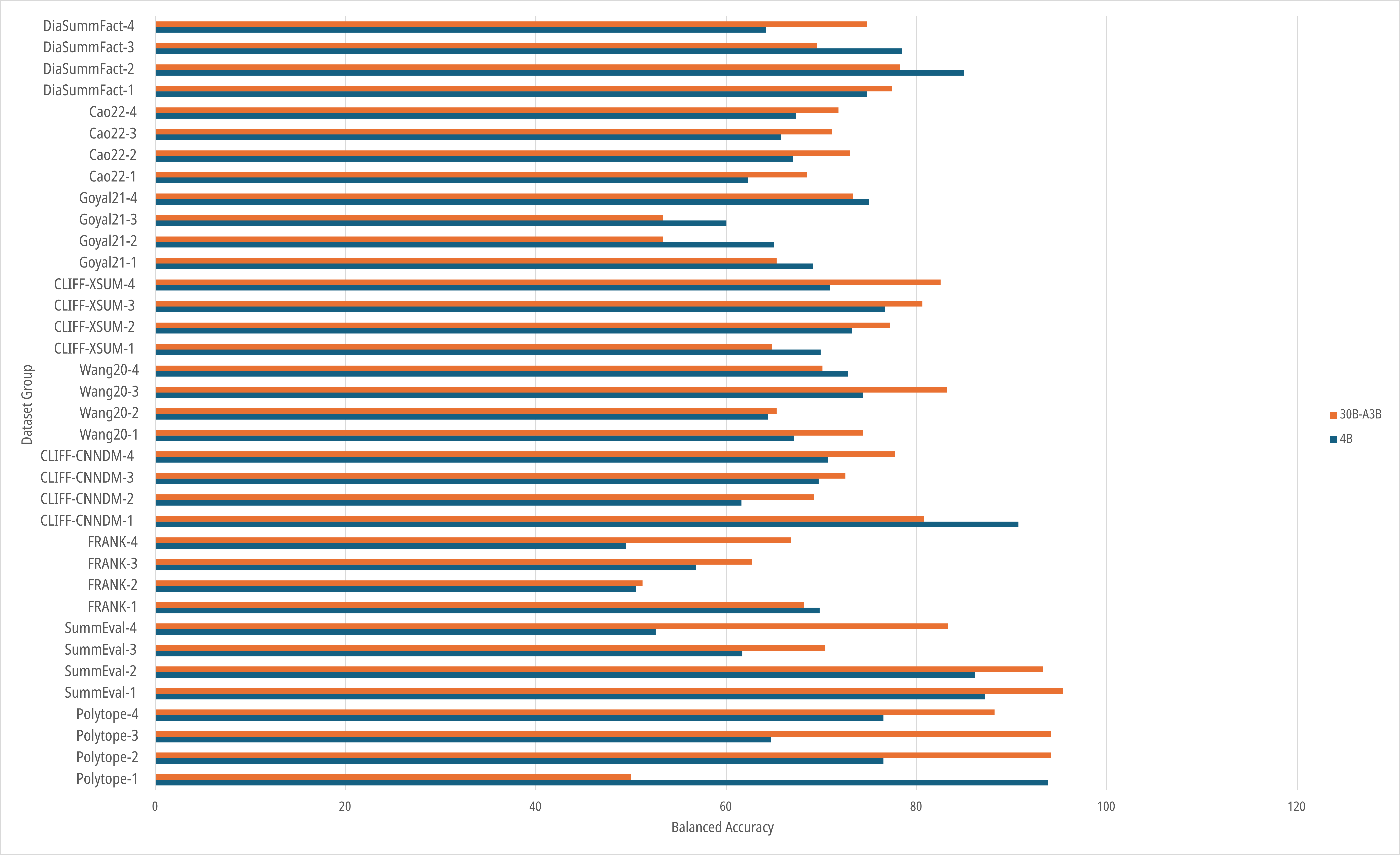}
    \caption{Balanced accuracies of dataset grouped by their context length on Factax-Benchmark.}
    \label{fig:long_context_factax}
\end{figure*}

\begin{figure*}
    \centering
    \includegraphics[width=\linewidth]{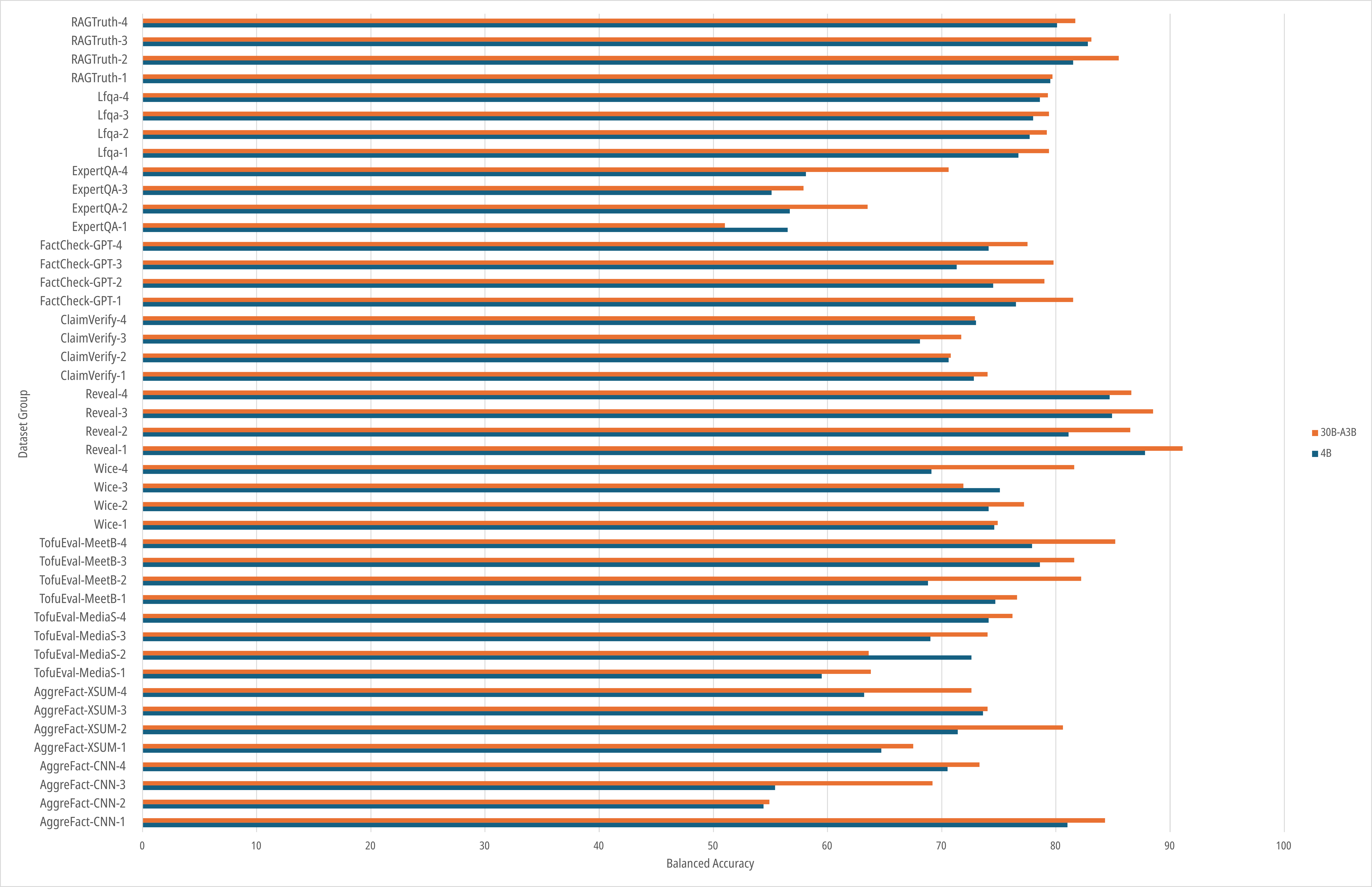}
    \caption{Balanced accuracies of dataset grouped by their context length on LLM-AggreFact.}
    \label{fig:long_context_LA}
\end{figure*}

To further investigate the impact of context length, we partitioned the evaluation data by source document length and computed balanced accuracy within each group, as shown in Figures \ref{fig:long_context_factax} and \ref{fig:long_context_LA}. 
We evenly divided the data in each subset into four groups based on document length, such that Group 1 < Group 2 < Group 3 < Group 4 in terms of context length.
We observe that the lowest performance does not consistently occur in the longest-document group. 
This pattern indicates that performance is more strongly influenced by the semantic complexity of the (document, claim) pair than by context length alone.

\newpage
\section{Impact of Threshold Tuning}
\label{app:threshold_tuning}
Our decision to use a fixed threshold of 0.5 is motivated by several considerations.

\begin{itemize}
    \item Compliance with benchmark protocol. The LLM-AggreFact \citep{tang-etal-2024-minicheck} requires metrics to apply the threshold at 0.5, so we follow this setting to ensure direct comparison with previously reported results and other metrics evaluated on the same benchmark.
    \item Limited impact on conclusions. Among metrics in our Table 1 and Table 2, only MiniCheck and FactCG will be affected by threshold selection, whereas our proposed metrics and other LLM-based metrics are not. Previous studies \citep{tang-etal-2024-minicheck,lei-etal-2025-factcg} have demonstrated the influence of dynamic thresholding on these two metrics. We show the results with and without dynamic thresholds in Table \ref{tab:threshold} below. The dynamic threshold does not affect the overall conclusions in our paper. 
\end{itemize}

\begin{table}[!h]
    \centering
    \resizebox{0.95\columnwidth}{!}{
    \begin{tabular}{|c|l|c|cc|}
    \hline
     & \textbf{Metric} & \makecell{\textbf{Threshold}\\\textbf{Tuning}} & \textbf{FacTax-Benchmark} & \textbf{LLM-AggreFact} \\
    \hline
     \multirow{8}{*}{\rotatebox{90}{Baseline}} & \multirow{2}{*}{FactCG} & \faCheck & 71.0 & 77.2 \\
       & & \faTimes & 67.0 & 75.6 \\
    \cline{2-5}
      & \multirow{2}{*}{MiniCheck-DeBERTa} & \faCheck & 66.1 & 74.1 \\
       & & \faTimes & 62.1 & 73.1 \\
    \cline{2-5}
      & \multirow{2}{*}{MiniCheck-RoBERTa} & \faCheck & 61.3 & 74.4 \\
       & & \faTimes & 61.7 & 73.5 \\
    \cline{2-5}
      & \multirow{2}{*}{MiniCheck-FT5} & \faCheck & 74.8 & 75.5 \\
       & & \faTimes & 66.9 & 75.0 \\
    \hline
     \multirow{3}{*}{\rotatebox{90}{Ours}} & Qwen-30B-A3B-Instruct & \faTimes & 78.0 & 76.3 \\
       & Qwen-4B-Instruct& \faTimes & 73.0 & 75.6 \\
       & Qwen-0.6B (Trained) & \faTimes & 72.6 & 73.6\\
    \hline
    \end{tabular}
    }
    \caption{The impact of threshold tuning and the result comparison against our metrics.}
    \label{tab:threshold}
\end{table}

\section{Dataset Statistics}
\label{app:stat}
We present the statistics of the datasets involved in this paper.
\begin{table}[!h]
    \centering
    \resizebox{0.95\columnwidth}{!}{
    \begin{tabular}{lccccc}
    \toprule
     \textbf{Benchmark} & \textbf{Split} & \textbf{Size} & \textbf{\makecell[c]{Source\\Length}} & \textbf{\makecell[c]{Claim\\Length}} & \textbf{\makecell[c]{Consistent\\Ratio (\%)}} \\
    \midrule
     \multirow{2}{*}{FacTax-Benchmark} & dev & 1236 & 442.4 & 37.3 & 64.2 \\
     & test & 1592 & 395.3 & 43.5 & 49.4\\
    \midrule
     \multirow{2}{*}{LLM-AggreFact} & dev & 29320 & 580.8 & 22.2 & 75.7\\
     & test & 30420 & 559.4 & 22.5 &77.9 \\
    \bottomrule
    \end{tabular}
    }
    \caption{The statistics of the two benchmarks.}
    \label{tab:benchmarks}
\end{table}

\begin{table}
    \centering
    \resizebox{0.6\columnwidth}{!}{
    \begin{tabular}{cccc}
    \toprule
     \textbf{Split} & \textbf{Size} & \textbf{\makecell[c]{Claim\\Length}} & \textbf{\makecell[c]{Avg Facts\\per Claim}} \\
    \midrule
     training &  69948 & 23.4 & 3.0 \\
     test & 7250 & 23.7 & 3.0 \\
    \bottomrule
    \end{tabular}
    }
    \caption{The statistics of the curated dataset for claim decomposition learning.}
    \label{tab:de_stat}
\end{table}

\begin{table}
    \centering
    \resizebox{0.6\columnwidth}{!}{
    \begin{tabular}{cccc}
    \toprule
     \textbf{Split} & \textbf{Size} & \textbf{\makecell[c]{Reason\\Length}} & \textbf{\makecell[c]{Consistent\\Ratio (\%)}} \\
    \midrule
     training &  32461 & 897.9 & 33.7 \\
     test & 3608 & 1062.8 & 25.4 \\
    \bottomrule
    \end{tabular}
    }
    \caption{The statistics of the curated dataset for fact checking learning.}
    \label{tab:re_stat}
\end{table}

For the dataset type, FacTax has:
\begin{itemize}
    \item \textbf{News Summarisation} PolyTope \citep{huang-etal-2020-achieved}, SummEval \citep{fabbri-2021-summeval}, FRANK \citep{pagnoni-etal-2021-understanding}, CLIFF \citep{cao-wang-2021-cliff}, Wang20 \citep{wang-etal-2020-asking}, Cao22 \citep{cao-etal-2022-hallucinated}, Goyal21 \citep{goyal-durrett-2021-annotating}. These datasets are collectively known as AggreFact-SOTA
    \item \textbf{Dialogue Summarisation} DiaSumFact \citep{zhu-etal-2023-annotating}
\end{itemize}
LLM-AggreFact has:
\begin{itemize}
    \item \textbf{Summarisation} AggreFact-CNN/XSum \citep{tang-etal-2023-understanding}, TofuEval-MeetB/MediaS \citep{tang-etal-2024-tofueval}, RAGTruth \citep{niu-etal-2024-ragtruth}
    \item \textbf{Retrieval-augmented Generation} ClaimVerify \citep{liu-etal-2023-evaluating}, LFQA \citep{chen-2023-understanding}, ExpertQA \citep{malaviya-etal-2024-expertqa}, RAGTruth
    \item \textbf{Post-hoc Grounding} ExpertQA, REVEAL \citep{jacovi-etal-2024-chain}, FactCheck-GPT \citep{wang-etal-2024-factcheck}
    \item \textbf{Human Written Claims} WiCE \citep{kamoi-etal-2023-wice}
\end{itemize}

\newpage
\section{Prompts}
\label{app:prompt}
We list all the prompts involved in the experiments in this section.

\begin{table}[!h]
\tiny
\centering
\begin{tabularx}{\columnwidth}{X}
\toprule
Following the example below, segment the given claim into atomic facts only based on the claim itself. Output each fact with "-" as the start.\\
\\
Claim:\\
The parkway was opened in 2001 after just under a year of construction and almost two decades of community requests.\\
Facts:\\
- The parkway was opened in 2001.\\
- The parkway was opened after just under a year of construction.\\
- The parkway was opened after two decades of community requests.\\
\\
CLAIM:\\
<claim>\\
Facts:\\
\bottomrule
\end{tabularx}
\caption{The prompt for claim decomposition.}
\label{tab:de_prompt}
\end{table}

\begin{table}[h!]
\tiny
\centering
\begin{tabularx}{\columnwidth}{X}
\toprule
Read the article given below and answer the questions.\\
\\
ARTICLE:\\
<article>\\
\\
Here is a claim, answer the following questions. Please reason step by step, and output your final answer by "Final Answer: yes" or "Final Answer: no".\\
\\
CLAIM:\\
<claim>\\
\\
1) For the claim, are the object and the subject mentioned?\\
\\
2) If the object and the subject are mentioned, is their related information verifiable according to the article? If there is information not mentioned, carry it into the next question. If verifiable but incorrect, stop here and answer "Final Answer: no".\\
\\
3) Look at the relationships between the object and the subject, is their relationship mentioned? If not, can the relationship be inferred from the article? If the relationship stands, can the previous information not mentioned be inferred from the article?\\
\bottomrule
\end{tabularx}
\caption{The prompt for fact checking. Only in ablation study, \textit{<claim>} refers to a original claim. Otherwise, it is an atomic fact decomposed from the claim.}
\label{tab:re_prompt}
\end{table}

\begin{table}[!h]
\tiny
\centering
\begin{tabularx}{\columnwidth}{X}
\toprule
Read the article given below and answer the questions.\\
\\
ARTICLE:\\
<article>\\
\\
Read the claim given below, judge if the claim is supported by the article or not. Please reason step by step, and output your final answer by "Final Answer: yes" or "Final Answer: no".\\
\\
CLAIM:\\
<claim>\\
\bottomrule
\end{tabularx}
\caption{The direct prompt used for ablation study.}
\label{tab:vanilla_prompt}
\end{table}

\begin{table}[!h]
\tiny
\centering
\begin{tabularx}{\columnwidth}{X}
\toprule
You are a proficient writer that always help people to change text meaning.\\
\\
I will give you a piece of text. Please [write a new similar sentence with different subjects or objects/tweak or add details in the text/keep the subject and object the same, but change the predicates in the text or negate the verb]. Only output one changed text without any explanation.\\
\\
Text: \\
<TEXT>\\
\\
Rewritten text: \\
\bottomrule
\end{tabularx}
\caption{The prompt used for alter the claims. Different prompts are applied to make the claim violate different criteria.}
\label{tab:data_prompt}
\end{table}

\begin{table}[!h]
\tiny
\centering
\begin{tabularx}{\columnwidth}{X}
\toprule
You need to verify if the claim is supported by the article. You should perform this task in two steps.\\

1 ) Claim Decomposition\\

Follwing the example below, segment the given claim into atomic facts only based on the claim itself first. Output each fact with "-" as the start.\\

Example Claim:\\

The parkway was opened in 2001 after just under a year of construction and almost two decades of community requests.\\

Example Facts:\\

The parkway was opened in 2001.\\
The parkway was opened after just under a year of construction.\\
The parkway was opened after two decades of community requests.\\
Claim to decompose:\\

<claim>\\

Fact Verification\\

2 ) Read the article given below. For each fact you just obtained, answer the questions given below. Please reason step by step. If all facts are supported by the article, output your final answer by "Final Answer: yes", otherwise "Final Answer: no".\\

ARTICLE:\\

<article>\\

Questions:\\

For the claim, are the object and the subject mentioned?\\

If the object and the subject are mentioned, is their related information verifiable according to the article? If there is information not mentioned, carry it into the next question. If verifiable but incorrect, stop here and answer "Final Answer: no".\\

Look at the relationships between the object and the subject, is their relationship mentioned? If not, can the relationship be inferred from the article? If the relationship stands, can the previous information not mentioned be inferred from the article?\\
\bottomrule
\end{tabularx}
\caption{The prompt used for single-pass fact-checking.}
\label{tab:single_pass_prompt}
\end{table}

\begin{table}[!h]
\tiny
\centering
\begin{tabularx}{\columnwidth}{X}
\toprule
Read the article given below and answer the questions.\\

ARTICLE:\\

<article>\\

Here is a claim, answer the following questions. Please reason step by step, and output your final answer by "Final Answer: yes" or "Final Answer: no".\\

CLAIM:\\

<claim>\\

1) Does the claim explicitly mention both the subject and the object?\\

2) If both are mentioned, can the information connecting them be verified using the article? If some details are missing, carry those forward to the next step. If the information is verifiable but incorrect, stop and respond with “Final Answer: no.”\\

3) Examine the relationship between the subject and the object. Is this relationship stated in the article? If not, can it reasonably be inferred? If the relationship is valid, determine whether any previously missing information can also be inferred from the article.\\
\bottomrule
\end{tabularx}
\caption{The fact-checking prompt paraphrased by ChatGPT.}
\label{tab:chatgpt_prompt}
\end{table}

\begin{table}[!h]
\tiny
\centering
\begin{tabularx}{\columnwidth}{X}
\toprule
Read the article given below and answer the questions.\\

ARTICLE:\\

<article>\\

Here is a claim, answer the following questions. Please reason step by step, and output your final answer by "Final Answer: yes" or "Final Answer: no".\\

CLAIM:\\

<claim>\\

1) Does the claim explicitly identify both the subject and the object?\\

2) If both entities are present, is the provided detail supported by the text? If the data is present but contradicts the source, mark as "no" and terminate. If details are missing but not contradicted, proceed to the next step.\\

3) Is the connection between the entities stated or clearly implied? If a relationship exists, determine if any previously missing details can be logically deduced from the source material.\\
\bottomrule
\end{tabularx}
\caption{The fact-checking prompt paraphrased by Gemini.}
\label{tab:gemini_prompt}
\end{table}

\section{Qualitative Study Example}

\begin{table*}
    \tiny
    \centering
    \begin{tabularx}{2\columnwidth}{|X|}
    \hline
     \textbf{Prompt} \\
     \\
     Read the article given below and answer the questions.\\
     \\
     ARTICLE:\\
     \\
     The adapted artwork for the Sgt Pepper's Lonely Hearts Club Band album is worth about £70,000... The artwork is \textcolor{blue}{one of five Beatles album sleeves} to feature in the top 10, compiled by Record Collector magazine. \underline{\textcolor{blue}{Sir Peter Blake}'s original Sgt Pepper collage was doctored to replace the faces of The Beatles and other notable figures with those of} \underline{executives from the band's US label Capitol}... In second place, valued at £7,000, are the... \\
     Here is a claim, answer the following questions. Please reason step by step, and output your final answer by "Final Answer: yes" or "Final Answer: no".\\
     \\
     CLAIM: \\
     \\
     A beatles album sleeve designed by sir peter blake has topped a list of the most valuable rare records.\\
     \\
     1) For the claim, are the object and the subject mentioned?\\
     \\
     2) If the object and the subject are mentioned, is their related information verifiable according to the article? If there is information not mentioned, carry it into the next question. If verifiable but incorrect, stop here and answer "Final Answer: no".\\
     \\
     3) Look at the relationships between the object and the subject, is their relationship mentioned? If not, can the relationship be inferred from the article? If the relationship stands, can the previous information not mentioned be inferred from the article?\\
     \\
    \hline
     \textbf{Facts from the LLM} \\
     - \textcolor{blue}{A Beatles album sleeve} was designed by \textcolor{blue}{Sir Peter Blake}.\\
     - The Beatles album sleeve topped a list of the most valuable rare records.\\
    \hline
     \textbf{Facts from the trained SLM} \\
     - \textcolor{blue}{A Beatles album sleeve} was designed by \textcolor{blue}{Sir Peter Blake}. \\
     - The Beatles album sleeve has topped a list of the most valuable rare records.\\
    \hline
     \textbf{Checking Fact 1 by LLM} \\
     \\
     1) Yes, both the object and the subject are mentioned.  \\
     - The subject is ""\textcolor{blue}{Sir Peter Blake}"".  \\
     - The object is ""\textcolor{blue}{a Beatles album sleeve}"".  \\
     \\
     2) Yes, the related information is verifiable according to the article.  \\
     - The article states: ""\underline{Sir Peter Blake's original Sgt Pepper collage was doctored...}""  \\
     - This confirms that Sir Peter Blake created the original artwork for the Sgt Pepper's Lonely Hearts Club Band album, which is a Beatles album sleeve.  \\
     - Therefore, the connection between Sir Peter Blake and a Beatles album sleeve is supported.\\
     \\
     3) Yes, the relationship between the subject and the object is directly mentioned in the article.  \\
     - The article explicitly states that Sir Peter Blake created the original artwork for the Sgt Pepper's Lonely Hearts Club Band album.  \\
     - This confirms the relationship: Sir Peter Blake designed the album sleeve.  \\
     - No additional information needs to be inferred, as it is clearly stated.\\
     \\
     Final Answer: yes \\
    \hline
    \textbf{Checking Fact 1 by the trained SLM} \\
    \\
    1) Yes, both the object (\textcolor{blue}{a Beatles album sleeve}) and the subject (\textcolor{blue}{Sir Peter Blake}) are mentioned in the article.\\
    \\
    2) Yes, the information is verifiable according to the article. The article states: ""\underline{Sir Peter Blake's original Sgt Pepper collage was doctored to replace the faces of The Beatles and} \underline{other notable figures with those of executives from the band's US label Capitol.}"" This confirms that a Beatles album sleeve was designed by Sir Peter Blake.\\
    \\
    3) The relationship between the object (a Beatles album sleeve) and the subject (Sir Peter Blake) is explicitly mentioned in the article — Sir Peter Blake designed the original Sgt Pepper collage, which was a Beatles album sleeve. This relationship is clearly stated and supported.\\
    \\
    Final Answer: yes\\
    \hline
    \end{tabularx}
    \caption{An example from the benchmark. The pinpointed information in the claim and the source is highlighted in \textcolor{blue}{blue}. The evidence sentence located is \underline{underlined}.}
    \label{tab:quality}
\end{table*}

\begin{table*}
\tiny
\centering
\begin{tabularx}{2\columnwidth}{|X|}
\hline
\textbf{Evidence sentence}: More than 1,700 people are employed in the division, which has factories in Rotherham and Stocksbridge.\\
\\
\textbf{Fact to verify}: The Speciality Steels division is based in Rotherham\\
\\
\textbf{Rationale output}: 1) The subject is ""The Speciality Steels division"" and the object is ""Rotherham"". Both are mentioned in the article.\\
\\
2) The article states that the division has factories in Rotherham and Stocksbridge. Therefore, the presence of a factory in Rotherham is verifiable and mentioned. However, the article does not explicitly state that the division itself is ""based"" in Rotherham—only that it has a factory there. The word ""based"" could imply headquarters or primary operations, which is not confirmed in the text.\\
\\
3) The relationship between the Speciality Steels division and Rotherham is that the division has a factory in Rotherham. However, the claim says the division “is based in Rotherham,” which implies a primary or organizational base, not just a factory location. While it is possible to infer that Rotherham is a key location for the division, it cannot be definitively concluded from the article that the division is based there—only that it operates there. Since the article does not confirm the division's base location, the relationship in the claim cannot be fully verified based on the text.\\
\\
Final Answer: no\\
\hline
\end{tabularx}
\caption{Failure examples of the inference rules.}
\label{tab:rebuttal_example}
\end{table*}

\end{document}